\documentclass[10pt,twocolumn,letterpaper]{article}

\usepackage[pagenumbers]{cvpr} 

\usepackage{graphicx}
\usepackage{tikz}
\usetikzlibrary{positioning}

\usepackage[pagebackref,breaklinks,colorlinks,citecolor=cvprblue]{hyperref}

\usepackage{xcolor}
\definecolor{cvprblue}{RGB}{0,113,188}

\def\paperID{5837} 
\def\confName{CVPR}
\def\confYear{2024}

\title{Fine-grained Appearance Transfer with Diffusion Models}

\author{
    YuTeng Ye$^1$, Guanwen Li$^1$, Hang Zhou$^1$, Jiale Cai$^1$, Junqing Yu$^1$, \\
    Yawei Luo$^2$, Zikai Song$^1$, Qilong Xing$^1$, Youjia Zhang$^1$, Wei Yang$^1$ \\
    $^1$Huazhong University of Science \& Technology, \\
    $^2$Zhejiang University \\
}

\let\oldtwocolumn\twocolumn
\renewcommand\twocolumn[1][]{%
    \oldtwocolumn[{#1}{
    
\begin{center}
\vspace{-1.5em}
\includegraphics[width=0.98\textwidth]{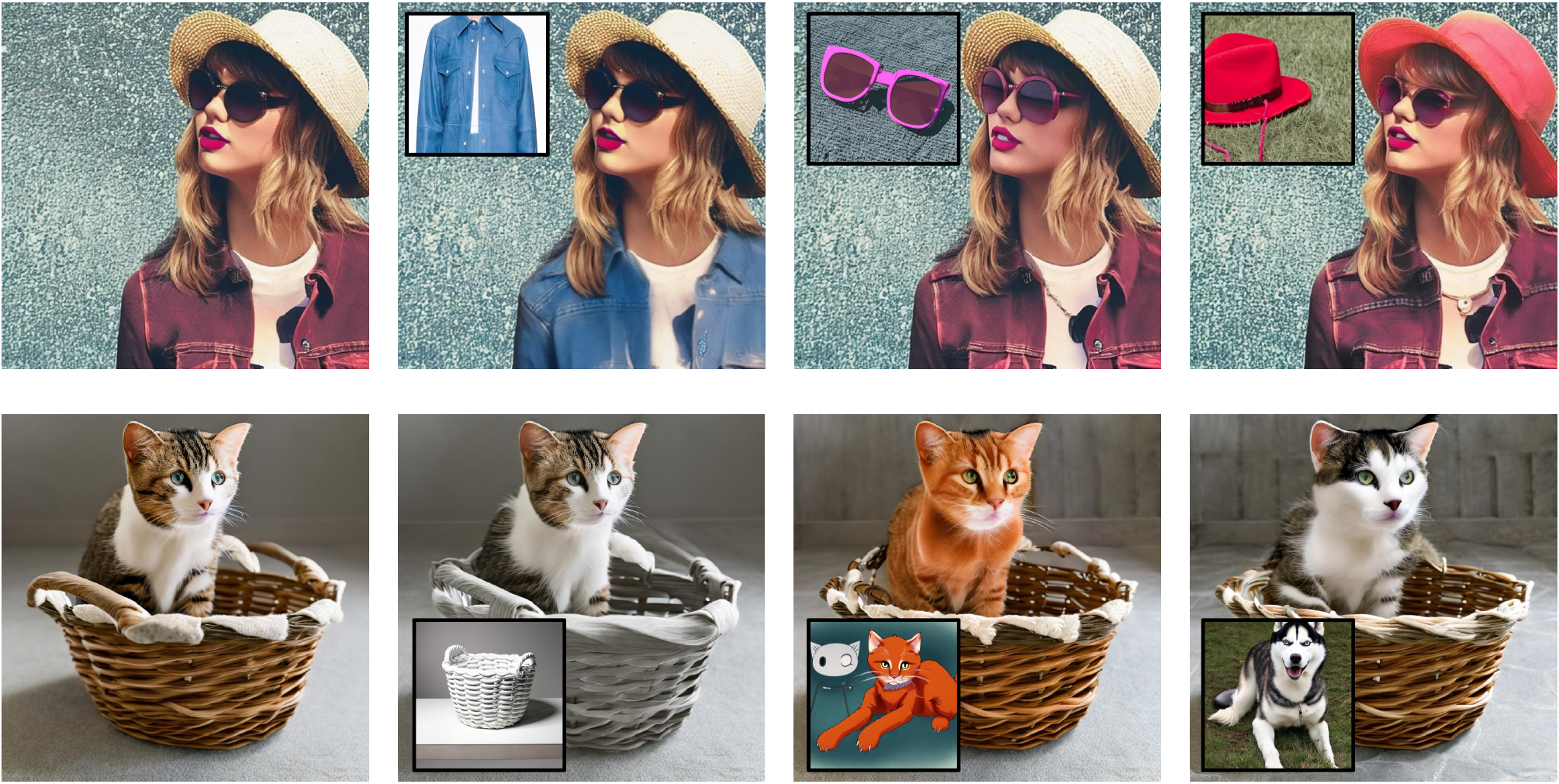}
    \captionof{figure}{The results of fine-grained appearance transfer using our method. The leftmost column displays the source images. On the right, the output images achieved by detailed appearance transfer corresponding to the target images (outlined in black), while preserving structural integrity. The examples at the bottom demonstrate our method's capability to transfer across various domains and categories.
    }
    \label{fig:teaser}
\end{center}
    }]
}

\begin{document}
\maketitle

\begin{abstract}
Image-to-image translation (I2I), and particularly its subfield of appearance transfer, which seeks to alter the visual appearance between images while maintaining structural coherence, presents formidable challenges. Despite significant advancements brought by diffusion models, achieving fine-grained transfer remains complex, particularly in terms of retaining detailed structural elements and ensuring information fidelity. This paper proposes an innovative framework designed to surmount these challenges by integrating various aspects of semantic matching, appearance transfer, and latent deviation.
A pivotal aspect of our approach is the strategic use of the predicted $x_0$ space by diffusion models within the latent space of diffusion processes. This is identified as a crucial element for the precise and natural transfer of fine-grained details. Our framework exploits this space to accomplish semantic alignment between source and target images, facilitating mask-wise appearance transfer for improved feature acquisition. A significant advancement of our method is the seamless integration of these features into the latent space, enabling more nuanced latent deviations without necessitating extensive model retraining or fine-tuning.
The effectiveness of our approach is demonstrated through extensive experiments, which showcase its ability to adeptly handle fine-grained appearance transfers across a wide range of categories and domains. We provide our code at https://github.com/babahui/Fine-grained-Appearance-Transfer

\end{abstract}
    
\section{Introduction}
\label{sec:intro}
Image-to-image translation (I2I) stands as a cornerstone task in computer vision, which primarily focuses on transforming images from a source domain to a corresponding target domain. Within the I2I spectrum, a notable sub-task is appearance transfer. The primary objective here is to transpose the visual aesthetics of a target image onto a source image, concurrently preserving the structural essence of the source image. Nonetheless, this task often grapples with complexities in achieving fine-grained transfer, especially in retaining intricate structures and upholding information fidelity.

The advent of diffusion models has catalyzed remarkable progress in the field, propelled by the emergence of large-scale models~\cite{saharia2022photorealistic,rombach2022high,ramesh2022hierarchical} and expansive datasets~\cite{sharma2018conceptual,schuhmann2022laion}. These models have risen to prominence due to their efficacy in various image translation tasks. However, their application in fine-grained appearance transfer encounters significant hurdles, primarily in two areas:

\begin{itemize}
    \item \textbf{Representation of Fine-Grained Information:} While current methods~\cite{kawar2023imagic,hertz2022prompt,chefer2023attend} often rely on text-based guidance for image generation within diffusion models, achieving precision in fine-grained control remains an elusive goal. This is attributed to the inherent challenges in pinpointing specific details and processing nuanced information. Conversely, image-guided methods offer a more direct approach by leveraging the abundant structural and semantic details intrinsic to images. However, some strategies, such as encoding the target image into low-level features for guiding generation~\cite{yang2023paint,tumanyan2022splicing}, can inadvertently oversimplify the visual appearance, thus losing critical fine-grained details.

    \item \textbf{Fine-Grained Appearance Transfer:} Methods employing contrastive learning based on structural and appearance similarities~\cite{tumanyan2022splicing,kwon2023improving} often face the challenge of losing semantic and structural nuances during the transfer process. Recent developments in diffusion features~\cite{tang2023emergent} have demonstrated potent capabilities in establishing precise point-to-point correspondence, aiding in meticulous matching and transfer. SD-DINO~\cite{zhang2023tale}, for instance, uses pixel-level swapping based on these correspondences for an initial coarse transfer, but requires additional refinement and lacks adaptability across diverse domains and categories.
\end{itemize}

In response to these challenges, we propose a novel framework integrating Semantic Matching, Appearance Transfer, and Latent Deviation components. Our innovative approach hinges on the exploitation of the \( x_0 \) space, an integral data-informed component extracted from the latent space of diffusion models, identified as key to facilitating natural and precise fine-grained information transfer.
Specifically, our method involves establishing a semantic alignment between source and target images within the \( x_0 \) space, leveraging the robust semantic relationships inherent in diffusion features. This alignment enables mask-wise appearance transfer to attain the desired feature. The transferred feature is then smoothly integrated into the latent space, undergoing a linear refinement of the intermediate latent representation and the generation of adaptive noise. This novel approach ensures a seamless transition from the \( x_0 \) space to the latent space, negating the necessity for extensive model training or fine-tuning.
Through the aforementioned process, we devise a modified latent trajectory that gradually transitions towards the target domain.
 
Our contributions are delineated as follows:

\begin{itemize}
    \item We identify the \( x_0 \) space as optimal for appearance transfers and implement semantic matching to capture fine-grained details effectively.

    \item We engineer a fluid transition from the \( x_0 \) space to the latent space, obviating the need for model retraining or extensive fine-tuning.

    \item Our empirical evaluations validate the efficacy of fine-grained appearance transfers across diverse categories and domains, underscoring the method's versatility and robustness.
\end{itemize}

\section{Related Work}
\label{sec:Related Work}

Our work aims at fine-grained appearance transfer, representing a challenging sub-problem in the realm of Image-to-Image Translation.

\subsection{Image-to-Image Translation}
Image-to-Image Translation (I2I) involves transforming images from a source domain to a target domain. Previous research~\cite{liu2021smoothing,choi2018stargan,wu2019relgan,isola2017image,gabbay2021scaling,huang2018multimodal,yi2017dualgan} primarily focused on models based on GANs~\cite{goodfellow2014generative}, yet adapting these models to diverse image domains has been challenging~\cite{su2022dual}.
The advent of diffusion models~\cite{ramesh2022hierarchical,rombach2022high} has brought significant advancements, offering enhanced flexibility and broader applicability in various contexts~\cite{kwon2022diffusion,cheng2023general}.
A specific subset of I2I is fine-grained appearance transfer. This subdomain concentrates on transferring visual appearance with an emphasis on preserving structural details and fidelity. While several studies~\cite{matsunaga2022fine,cheng2023general} have addressed fine-grained transfer, consistently achieving high-quality results in a wide range of scenarios remains a substantial challenge. Our work is dedicated to this aspect, seeking to adapt fine-grained appearance transfer for broad applicability across diverse categories and domains.

\subsection{Predicted $x_0$ Feature in Diffusion Model}
In the denoising process of diffusion models, noisy latent features are projected to an estimated \( x_0 \) feature, which encapsulates clean information. Several research methodologies have been developed based on this feature. For example, SAG~\cite{hong2023improving} utilizes Gaussian blur to reduce redundant information in \( x_0 \) feature, thereby enhancing sample quality. Additionally, General I2I~\cite{cheng2023general} focuses on minimizing the distance between \( x_0 \) features to reconstruct the original image. Other methods~\cite{kwon2023improving} employ an image encoder to process the \( x_0 \) feature and then transfer it with structural or appearance constraints.
In contrast to these approaches, our primary goal is to facilitate the transfer of fine-grained information at the \( x_0 \) feature level. Our approach differs from methods like those in~\cite{kwon2023improving} that encode the \( x_0 \) feature through a model; instead, we engage in fine-grained semantic matching directly within the \( x_0 \) feature. This approach ensures the preservation of detailed structural integrity and information fidelity.

\subsection{Feature Correspondence}
Feature correspondence in generative models establishes dense connections among features derived from a range of visual appearances and categories. Initially, methods like SIFT~\cite{david2004distinctive} and HOG~\cite{dalal2005histograms} relied on hand-designed features. The introduction of deep generative models such as GANs~\cite{goodfellow2020generative} represented a shift from these traditional approaches, contributing to the identification of visual correspondences in diverse image domains~\cite{peebles2022gan,mu2022coordgan,zhang2021datasetgan}.
Recent advances, exemplified by DIFT~\cite{tang2023emergent}, have shown that diffusion models' intermediate features exhibit robust point-to-point correspondence. Utilizing this feature, various studies have applied feature correspondence to downstream tasks. DIFT~\cite{tang2023emergent} uses these correspondences for affine transformations from a target to a source image. DragonDiffusion~\cite{mou2023dragondiffusion} applies a feature correspondence loss to guide the editing of intermediate features. Additionally, SD-DINO~\cite{zhang2023tale} combines elements from DINO~\cite{caron2021emerging} and Stable Diffusion to compute feature correspondences for object swapping.
Contrasting with SD-DINO~\cite{zhang2023tale}, which focuses on swapping within the same categories, our work extends to a broader range of appearance transfer scenarios, including transfers across different domains and categories.

\section{Methods}
\subsection{Preliminaries}
Similar to DDPM \cite{ho2020denoising}, DDIM \cite{song2020denoising} reconstructs an image from white noise through an iterative denoising process. DDIM, aiming to accelerate the sampling process of DDPM while ensuring the quality of generated images, devises a process with fewer steps and a non-Markovian nature. Formally, given an image \( x_0 \) and a variance schedule \( \beta_t \) at timestep \( t \), the generation process in DDIM is deterministic. The latent \( x_t \) comprises the predicted \( x_0 \) and a component directed towards \( x_t \), which can be described by the following equation:
\begin{align}
    x_{t-1} & = \sqrt{\alpha_{t-1}} \underbrace{\left(\frac{x_t - \sqrt{\beta_{t}} \epsilon_\theta^{(t)}(x_t)}{\sqrt{\alpha_t}}\right)}_{\text{`` predicted } x_0 \text{''}} + \underbrace{\sqrt{\beta_{t-1}} \cdot \epsilon_\theta^{(t)}(x_t)}_{\text{``direction pointing to } x_t \text{''}} \label{eq:ddim}
\end{align}

where \( \alpha_t  = 1 - \beta_{t} \), and \( \epsilon_\theta^{(t)} \) denotes a neural network parameterized by \( \theta^{(t)} \). 
\
In~\cref{eq:ddim}, the latent variable representation \( x_t \) can be viewed as a blend of the data-informed component \( x^{(t)}_0 \) and the noise-driven term \( \epsilon(x_t) \), formalized as:
\begin{equation}
    x_t = \sqrt{a_t} x^{(t)}_0 + \sqrt{1-a_t} \epsilon_\theta^{(t)}(x_t) \label{eq:x0} 
\end{equation}
For brevity in subsequent sections, we denote the term \( \epsilon_\theta^{(t)}(x_t) \) as \( \epsilon(x_t) \), and \( x^{(t)}_0 \) as \( x_0 \).

\subsection{Problem Formulation}
%
Our proposed framework aims to facilitate a detailed, fine-grained synthesis between a source image \( I \) and a target image \( T \). This process entails the precise transfer of visual appearance from \( I \) to \( T \), with an emphasis on maintaining the structural integrity of \( I \). The primary challenges in this fine-grained transformation include:

\begin{itemize}
    \item \textbf{Detail Alignment}: Achieving accurate alignment of fine-grained details between \( I \) and \( T \), crucial for preserving the structural integrity of objects during the transformation process.
    \item \textbf{Fine-grained Appearance Transformation}: Implementing a transformation that meticulously aligns with semantic and structural nuances across different domains and categories, thus enabling a seamless and coherent integration of visual elements.
\end{itemize}

\begin{figure}[t]
    \centering
    \includegraphics[width=0.48\textwidth]{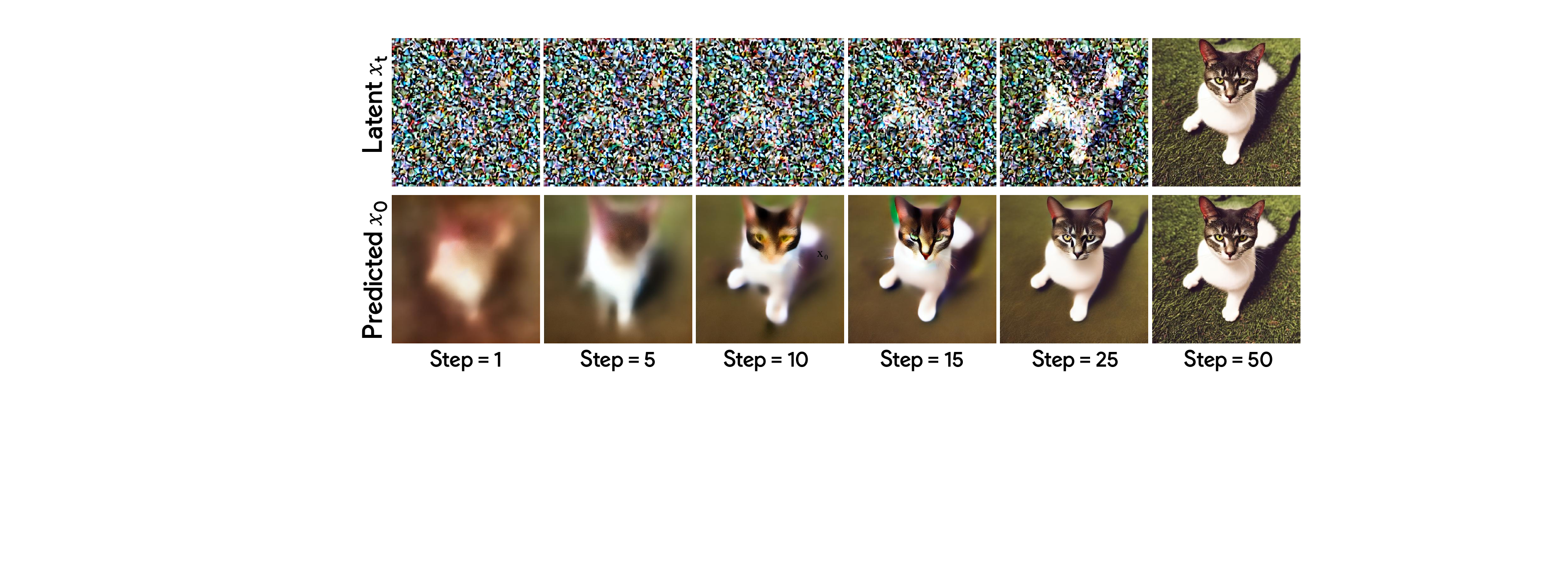}
    \caption{Illustration of DDIM Sampling Steps with Latent Feature $x_t$ and Predicted $x_0$. Here, we apply a latent decoder to convert both predicted $x_0$ and $x_t$ into the image for visualization.}
    \label{fig:Observations}
\end{figure}

\begin{figure*}[t]
    \centering
    \includegraphics[width=0.95\textwidth]{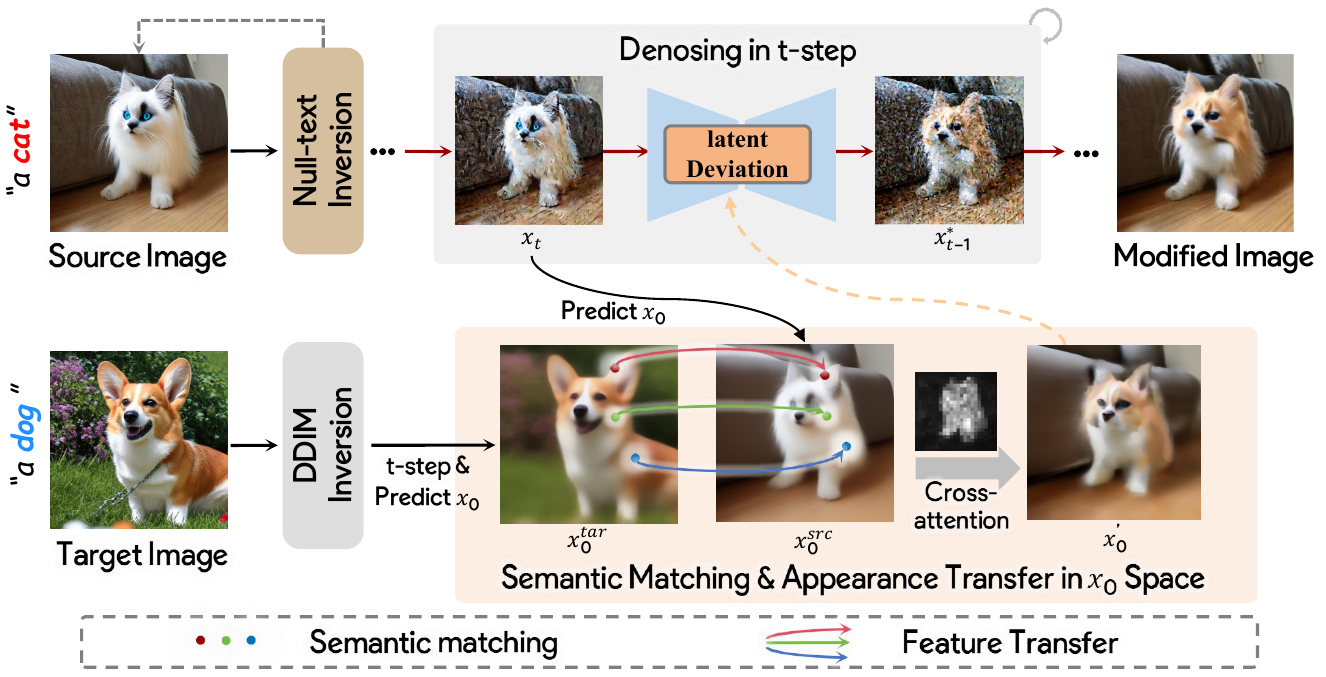}
    \caption{
    \textbf{Our Pipeline.} This figure illustrates our pipeline, commencing with null-text inversion applied to the source image \( I \), creating a latent path for reconstructing the image. During the diffusion denoising stage, Latent Deviation is performed, leading to a modified image that aligns with the target image \( T \). Specifically, the process begins with semantic alignment in the \( x_0 \) space between \( x_0^{\text{src}} \) and \( x_0^{\text{tar}} \), where \( x_0^{\text{tar}} \) is obtained through DDIM inversion with \( T \). Based on semantic relations, features from \( x_0^{\text{tar}} \) are transferred to \( x_0^{\text{src}} \), guided by an attention mask of \( I \), resulting in \( x'_0 \). Finally, \( x'_0 \) is processed in the latent space to synthesize the final modified image.
    }
    \label{fig:pipeline}
\end{figure*}

\begin{figure}[t]
    \centering
    \includegraphics[width=0.48\textwidth]{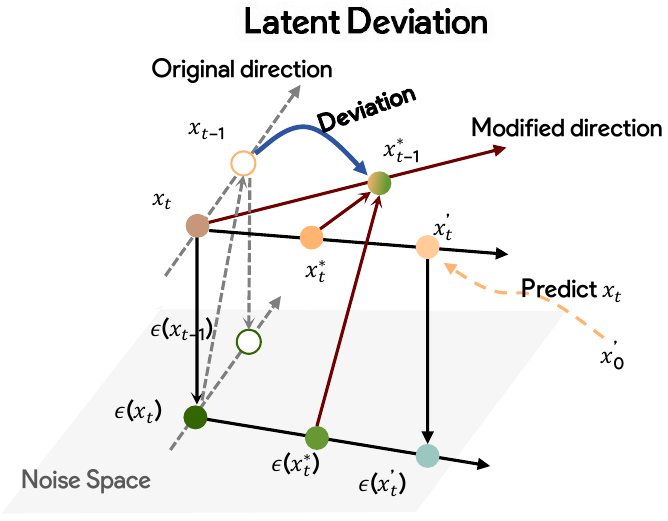}
    \caption{
    \textbf{Latent Deviation Process.} Dashed lines represent the original latent path for image reconstruction, while solid red lines indicate the modified latent trajectory. Initially, the transferred \( x'_0 \) from \cref{fig:pipeline} is transformed into its corresponding latent code \( x'_t \) and its noise component \( \epsilon(x'_t) \). Subsequently, a linear update refines the intermediate latent representation \( x^*_t \) and adjusts the noise term \( \epsilon(x^*_t) \). Finally, \( x^*_t \) and \( \epsilon(x^*_t) \) undergo a forward DDIM pass, leading to the next denoising step and producing \( x^*_{t-1} \). This step highlights the deviation from the original \( x_{t-1} \) to the modified \( x^*_{t-1} \) at step \( t \).
    }
    \label{fig:Latent Deviation}
\end{figure}

\subsection{Overall Framework}
As depicted in~\cref{fig:pipeline}, our methodology initiates by applying null-text inversion~\cite{mokady2023null} to the source image \( I \) along with the associated source prompt, optimizing unconditional text embedding at each step to create a latent path \(\{x_t, x_{t-1}, \ldots, x_0\}\) for reconstructing \( I \). Concurrently, DDIM inversion~\cite{song2020denoising} is applied to the target image \( T \) and its corresponding target prompt, progressively introducing noise to transition into the latent space. Throughout this process, we predict the \( t \)-step \( x_0^{\text{tar}} \) image feature from its latent code using~\cref{eq:x0}.

The framework integrates two pivotal concepts: Semantic Matching \& Appearance Transfer and Latent Deviation. These approaches allow us to alter the generation path of \( x_t \) using \( x_0^{\text{tar}} \), a strategy informed by our insights into the \( x_0 \) space. We observe a notable potential for manipulating image features in this space while maintaining structural integrity.

\textbf{Observations in \( x_0 \) Space.}
Our analysis, as shown in \cref{fig:Observations}, uncovers a key finding: the initial latent code \( x_0 \) exhibits a higher fidelity in capturing fine-scale details compared to the noisier \( x_t \). Particularly in the early stages of the diffusion process, \( x_0 \) plays a crucial role in reconstructing the object's structure, progressively evolving to encapsulate more complex details. We define the series of steps involving \( x_0 \) as the \( x_0 \) space, which emerges as a vital medium for information transfer between images. Transferring features in the \( x_0 \) space effectively minimizes perturbations induced by latent noise, thereby preserving and conveying high-fidelity details during image synthesis.

\textbf{Semantic Matching and Appearance Transfer.}
%
Our objective is to achieve a semantic-aligned visual appearance transfer from a source image to a target image. We propose utilizing the \( x_0 \) space for this transformation due to its capability to retain structured and fine-scale image features.
To facilitate this transfer within the \( x_0 \) space, establishing a semantic alignment between the source and target is crucial. We employ the DIFT Feature~\cite{tang2023emergent}, which demonstrates strong relations across different domains and categories, for semantic matching in the \( x_0 \) space.
Specifically, given a source image feature \( x_0^{\text{src}} \) and a target image feature \( x_0^{\text{tar}} \) at step \( t \), we input them into a diffusion model to extract their respective DIFT features. We then compute the point-to-point cosine distance between these features, ranking these distances to form a correlation mapping \( \mathcal{C} \) from source to target.
Lastly, we perform a mask-wise transfer in the \( x_0 \) space, targeting only the primary objects for transformation while preserving the background. This whole transfer approach in \( x_0 \) space is formulated as:
\begin{equation}
    x'_0 = \mathcal{M} \left( (1-\delta) x_0^{\text{src}} + \delta \mathcal{C}(x_0^{\text{tar}}) \right) + (1-\mathcal{M}) x_0^{\text{src}}
    \label{eq:transfer-x0}
\end{equation}
where \( x'_0 \) represents the transferred image feature, \( \mathcal{M} \) denotes a cross attention mask applied to primary objects in source image. \( \delta \) signifies the weight of the transfer, and \( \mathcal{C}(x_0^{\text{tar}}) \) represents the features of \( x_0^{\text{tar}} \) that are matched in \( x_0^{\text{src}} \).
We simplify the appearance transfer described in \cref{eq:transfer-x0} as follows:
\begin{equation}
    x'_0 = x_0 + \mathcal{T}(x'_0, x_0)
    \label{eq:simplify-transfer-x0}
\end{equation}
Here, \(\mathcal{T}(x'_0, x_0)\) denotes the transformation from \(x_0\) to \(x'_0\).

\textbf{Latent Deviation.}
The next stage, termed ``Latent Deviation,'' involves transitioning the transferred \( x'_0 \) into a latent space for smoother processing, as depicted in \cref{fig:Latent Deviation}. The transition begins by converting \( x'_0 \) to a latent code \( x'_t \) according to \cref{eq:x0,eq:simplify-transfer-x0}, which is expressed by the following equation:
\begin{equation}
    x'_t = \sqrt{a_t} (x_0 + \mathcal{T}(x'_0, x_0)) + \sqrt{1-a_t} \epsilon(x_t) 
\label{eq:transfer-xt}
\end{equation}
Here, we propagate appearance differences, represented by \( \mathcal{T}(x'_0, x_0) \), to \( x'_t \).
Following this, \( x'_t \) is processed through a U-Net to derive its noise component \( \epsilon(x'_t) \), indicating the current noise direction toward \( x'_t \). Instead of directly using \( x'_t \) for further processing, we compute a smoother latent representation, given by:
\begin{equation}
    x^*_t = \lambda x_t + (1-\lambda) x'_t
    \label{eq:update-latent}
\end{equation}
where \( \lambda \) represents the balance coefficient. The noise term \( \epsilon(x^*_t) \) is then updated as a linear mixture:
\begin{equation}
    \epsilon(x^*_t) = \gamma \epsilon(x_t) + (1-\gamma) \epsilon(x'_t)
    \label{eq:update-noise}
\end{equation}
In the denoising process, \( x^*_t \) and \( \epsilon(x^*_t) \), which are aligned with the target domain, are input into a subsequent forward pass through the DDIM. This results in obtaining \( x^*_{t-1} \) for the next denoising step. The whole latent deviation path in the denoising process is denoted as \( \{x_t, x^*_{t-1}, x^*_{t-2}, \ldots, x^*_{t-k}\} \).

\textbf{Framework Explanation.}
Our method smoothens the transfer of appearance differences from the \( x_0 \) space to latent deviations. To further analyze the denoising process from step \( t \) to \( t-1 \), we substitute \cref{eq:transfer-xt,eq:update-noise,eq:update-latent} into \cref{eq:ddim}. This yields the following refined formulation:
\begin{equation}
\begin{split}
    x^*_{t-1} = & \sqrt{\alpha_{t-1}} (x_0 + (1-\lambda) \mathcal{T}(x'_0, x_0)) \\
    & + \sqrt{1-a_{t-1}} \epsilon(x^*_t) \\
    & + \sqrt{\frac{a_{t-1}(1-a_t)}{a_t}} (\epsilon(x_t) - \epsilon(x^*_t))
\end{split}
\label{eq.6}
\end{equation}

Here, \( (1-\lambda) \mathcal{T}(x'_0, x_0) \) represents the shift within the \( x_0 \) space towards the target domain and the noise term \( \epsilon(x^*_t) \) serves as a construct directing towards \( x^*_t \). The term \( \epsilon(x_t) - \epsilon(x^*_t) \) is tailored to adapt to the denoising process.

\section{Experiment}
\textbf{Implementation details.} 
We conduct experiments on a single NVIDIA V100 GPU using Stable Diffusion v2~\cite{rombach2022high}. In the initial phase, we apply null-text inversion to fine-tune unconditional text embedding. Our sampling process then proceeds without requiring any further model training or fine-tuning.
For semantic matching, we employ DIFT matching~\cite{tang2023emergent} to obtain \( \mathcal{C} \) in \cref{eq:transfer-x0}. We set the parameter \( \delta \) in \cref{eq:transfer-x0} to 0.6, regulating the extent of feature transfer. Additionally, we configure \( \lambda \) in \cref{eq:update-latent} to 0.2, and \( \gamma \) in \cref{eq:update-noise}, which updates the unconditional noise component, is also set to 0.2. We set the start step to 12 to achieve a stable attention map. We set the end step at 21 based on observations of stable transfer achieved in the generated images.

\begin{figure*}[t]
    \centering
    \includegraphics[width=0.95\textwidth]{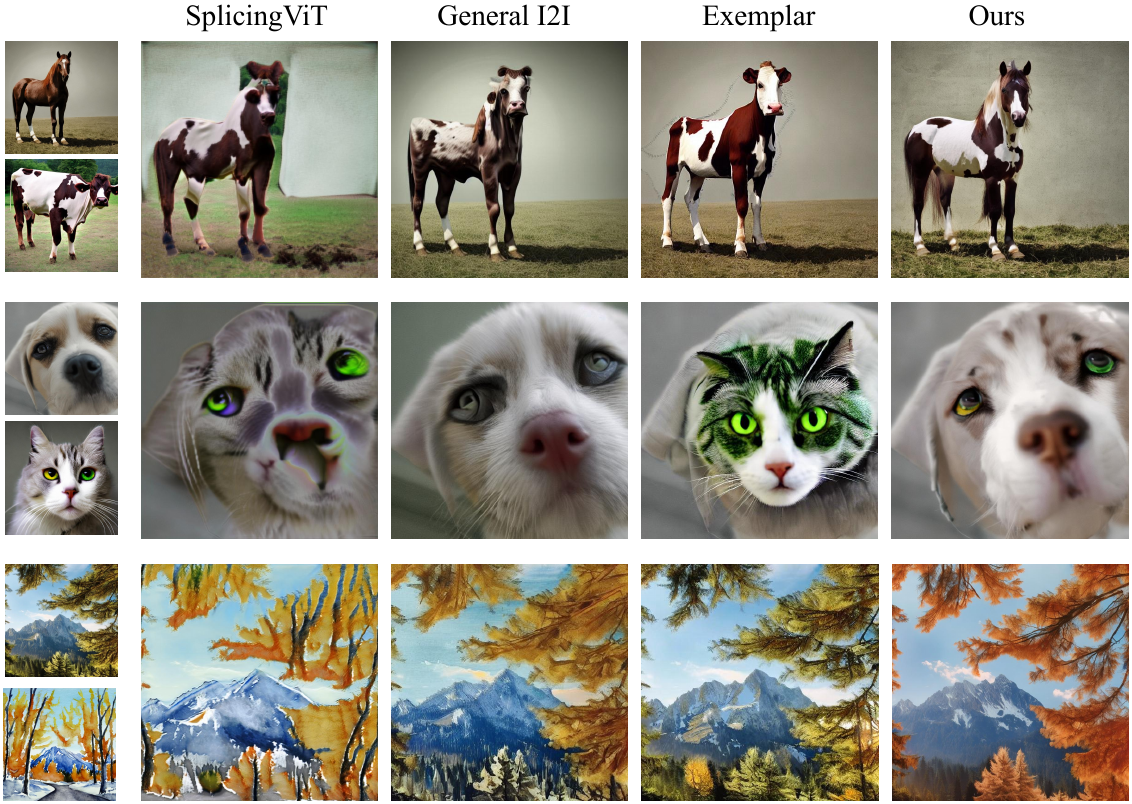}
    \caption{\textbf{Qualitative Comparisons with Image-Guided Transfer Methods.} The figure showcases appearance transformations in three sets, arranged from top to bottom: from ``horse'' to ``cow'', ``dog'' to ``cat'' in a face swap, and from a real photo to a painting. In each set, the leftmost images are the source (top) and target (bottom).}  
    \label{fig:image-guide figs}
\end{figure*}

\textbf{Datasets.}
For evaluating our method, we collect data from both synthetic and real-world sources, as there is no standard dataset specifically tailored to our task. We generate synthetic data in a two-step process. First, we generate source and target text pairs using online ChatGPT-4~\cite{OpenAI2023GPT4TR}, covering a semantic range from closely related concepts (\eg ``dog'' to ``Welsh Corgi'') to those with larger gaps (\eg ``dog'' to ``panda''), or from different domains (\eg ``a photo of a cat'' to ``an animation-style image of a cat''). We then process these text prompts through Stable Diffusion~\cite{rombach2022high} to produce images. From this process, we manually select 400 pairs of high-quality source and target images for our evaluations, ensuring a diverse representation of scenarios including landscapes and animals.
Additionally, we compile a set of 100 real-world images with related source and target text. These images are sourced and downloaded from the internet, adhering to permissible conditions.

\begin{table}[h]
    \centering
    \resizebox{\linewidth}{!}{%
    \begin{tabular}{cccc}
        \toprule
        \textbf{Method} & CLIPscore~$\uparrow$ & USERscore~$\uparrow$ & Total Cost Time \\
        \midrule
        SplicingViT~\cite{tumanyan2022splicing} & 0.273 & 34.8 & $\sim$1h \\
        \midrule
        General I2I~\cite{cheng2023general} & 0.289 & 64.5 & $\sim$30min \\
        \midrule
        Exemplar~\cite{yang2023paint} & 0.268 & 31.7 & - \\
        \midrule
        Ours & \textbf{0.301} & \textbf{75.1} & $\sim$3min \\
        \bottomrule
    \end{tabular}%
    }
    \caption{\textbf{Quantitative comparison with Image-guided Transfer Methods.}}
    \label{tab:1}
\end{table}

\subsection{Comparison to Current Work}
\textbf{Qualitative and Quantitative Comparisons with Image-guided Transfer Methods.} We compare our method with three current image-guided transfer approaches, including two diffusion models~\cite{cheng2023general,yang2023paint} and one ViT-based method~\cite{tumanyan2022splicing}.
SplicingViT~\cite{tumanyan2022splicing} primarily focuses on transferring semantic appearance, while General I2I~\cite{cheng2023general} addresses a wide range of image-to-image translation tasks, translating visual concepts across different domains. Exemplar~\cite{yang2023paint} incorporates an attention mechanism~\cite{mokady2023null} for generating necessary input masks.
Additionally, we did not include SD-DINO~\cite{zhang2023tale} in our comparison as it specializes in transferring identical objects and lacks comprehensive code availability.
The qualitative results are shown in~\cref{fig:image-guide figs}. SplicingViT tends to produce images with poor generation quality. General I2I retains the overall structure but struggles to effectively transfer fine details of appearance. As illustrated in the second row of~\cref{fig:image-guide figs}, crucial details like the target species' eye color are missed.
Exemplar achieves image-level appearance transfer but lacks detail alignment, evident in the first row of~\cref{fig:image-guide figs} where the color region on the cow's back is inaccurately transferred, possibly due to the use of image encoding.
Furthermore, these methods do not perfectly preserve detailed structure and information fidelity. For example, as shown in the first row of~\cref{fig:image-guide figs}, the shape of the horse's ears is not accurately maintained, and in the second row, the eye color is not effectively transferred.
Our result demonstrates the superiority of our method in achieving fine-grained appearance transfer.

\begin{figure*}[t]
    \centering
    \includegraphics[width=0.95\textwidth]{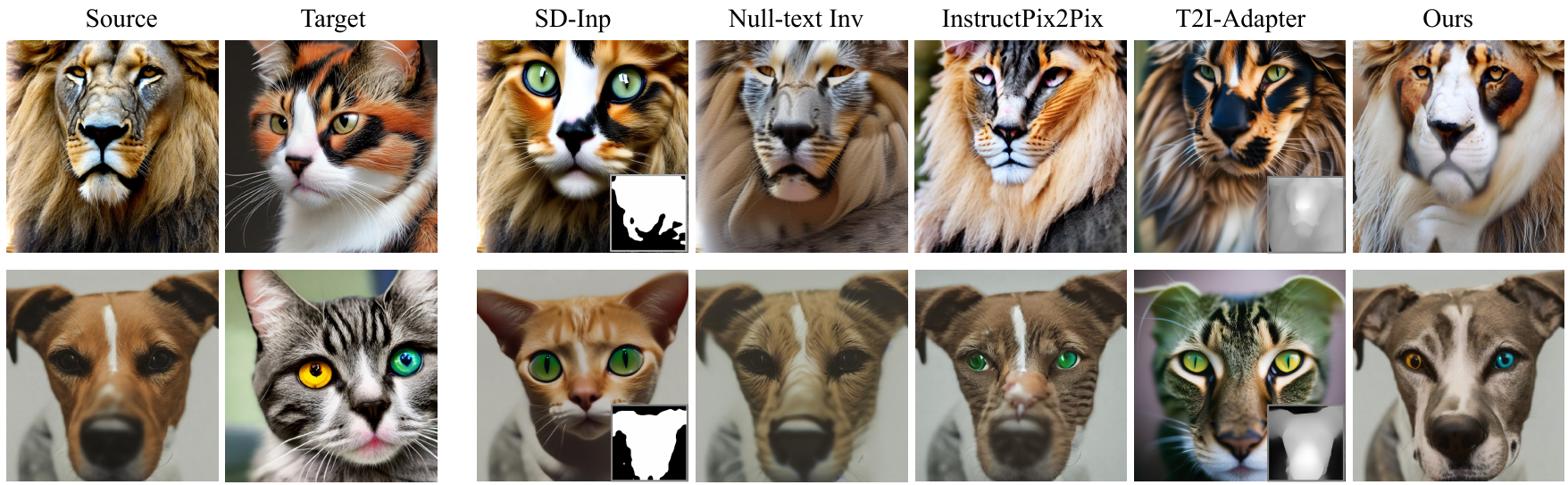}
    \caption{\textbf{Qualitative Comparisons with Text-Guided Transfer Methods.} For text input to text-guided methods, we generate detailed descriptions of target images using ChatGPT-4~\cite{OpenAI2023GPT4TR}.}
    \label{fig:text-guide figs}
\end{figure*}

\begin{figure}[t]
    \centering
    \includegraphics[width=0.48\textwidth]{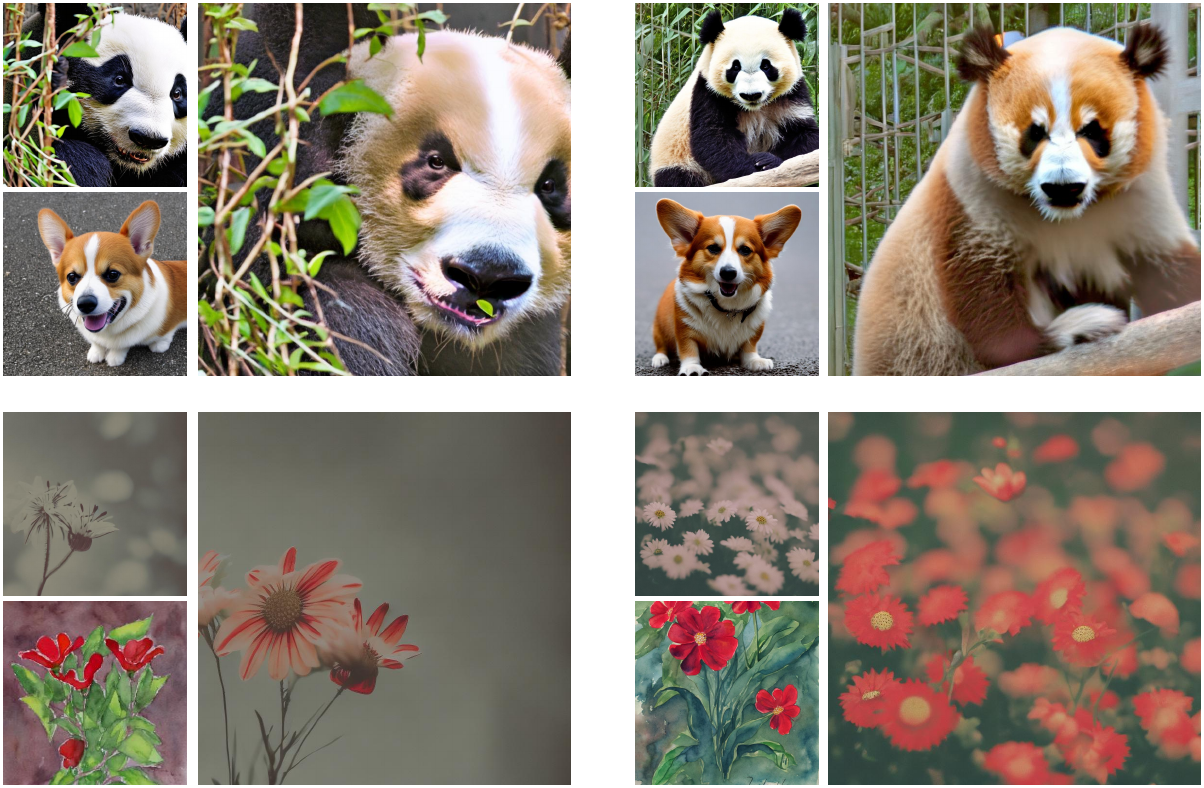}
    \caption{Results of our method in different domains and across large semantic gaps. The figure presents four sets of images: in each set, the top-left is the source image, the bottom-left is the target image, and the right side displays our output. The top two sets show the appearance transfer from ``a panda'' to ``a dog'', while the bottom two sets depict the transformation from realistic ``flowers'' to watercolor ``flower''.}
    \label{fig:res-cross-domain}
\end{figure}

We also conduct a quantitative evaluation with image-guided transfer methods. Due to the absence of ground truth for fine-grained appearance transfer, we assess the quality of appearance transfer from the following aspects:
1. To our knowledge, there is no dedicated metric specifically for measuring the quality of fine-grained transfer. Therefore, we conduct a user study where six human evaluators rate the transfer quality. Evaluators are asked to assess several aspects, including whether the overall appearance of the object is successfully transferred, whether the fine-grained transfer matches semantically, and whether the structural details are preserved. The user ratings range from 0 to 100, with higher scores indicating better quality of transfer. We calculate the average to determine the final score.
2. We utilize CLIP score~\cite{radford2021learning} to evaluate the semantic similarity between the source prompt and the output image, as our approach only involves appearance transfer, which is distinct from semantics.
Quantitative results, as shown in ~\cref{tab:1}, demonstrate that our method outperforms others in terms of CLIP score and user ratings. Regarding time efficiency, we compare the total sampling time, and our method is notably superior to others. The Exemplar method is not directly comparable because it requires approximately one week of model training in advance~\cite{yang2023paint}. The primary time cost of our method is attributed to the initial Null-text inversion~\cite{mokady2023null}, with subsequent sampling processes requiring no additional model fine-tuning or training.

\textbf{Qualitative Comparisons with Text-Guided Transfer Methods.} We further evaluate the capability of fine-grained appearance transfer by comparing with text-guided image translation methods, including Stable Diffusion Inpainting (SD-inp)~\cite{rombach2022high}, T2I-Adapter~\cite{mou2023t2i}, Null-text Inversion (Null-text Inv)~\cite{mokady2023null}, and InstructPix2Pix~\cite{brooks2023instructpix2pix}. For these methods, we first obtain detailed text descriptions of the target image using online ChatGPT-4~\cite{OpenAI2023GPT4TR}, which are then input into the models to guide the image modification.
The results are shown in \cref{fig:text-guide figs}. Our observations indicate that relying solely on text can be challenging for precisely controlling structural details and appearance. For example, as shown in the first row, using only text results in difficulties with accurately transferring specific facial structures, and in the second row, achieving exact eye color transfer through text alone is challenging.

To demonstrate the efficacy of our method across different domains and large semantic gaps, we present additional results in~\cref{fig:res-cross-domain}. As shown in the first row, our approach successfully transforms the appearance from ``a panda'' to ``a dog,'' notably retaining the dog's color and texture in the process. This exemplifies our method's effectiveness in aligning semantically disparate elements in scenarios with large semantic gaps.
The second row illustrates our capability to transfer appearances across different domains. 

\subsection{Ablation Study}

\begin{figure}[t]
    \centering
    \includegraphics[width=0.48\textwidth]{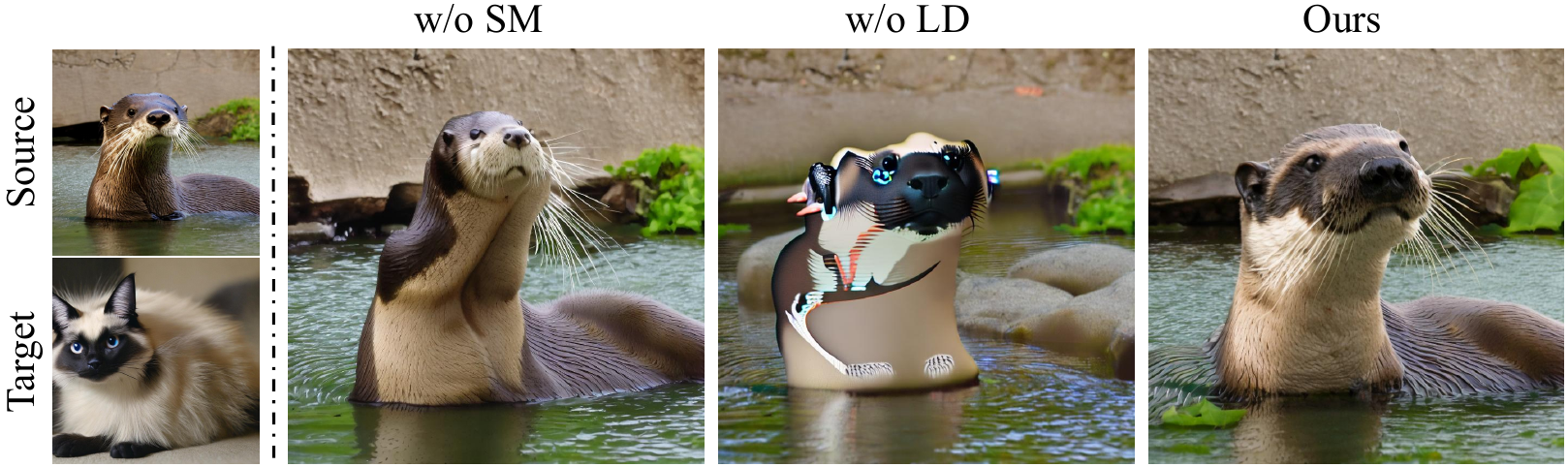}
    \caption{Visualization of the Ablation Study on Method Components.}
    \label{fig:ablation-module}
\end{figure}

\textbf{Ablation Study on Method Components.}
We conduct an ablation study on each component of our method, including Semantic Matching (SM) and Latent Deviation (LD). 
As shown in~\cref{fig:ablation-module}, without SM, our method fails to accurately match appearances, leading to incorrect information transfer. Additionally, without LD, the transfer process becomes uneven, resulting in implausible outcomes. Overall, by employing all of our proposed components, we achieve the best generation outputs, which better preserve both structure and appearance.

\begin{figure}[t]
    \centering
    \includegraphics[width=0.48\textwidth]{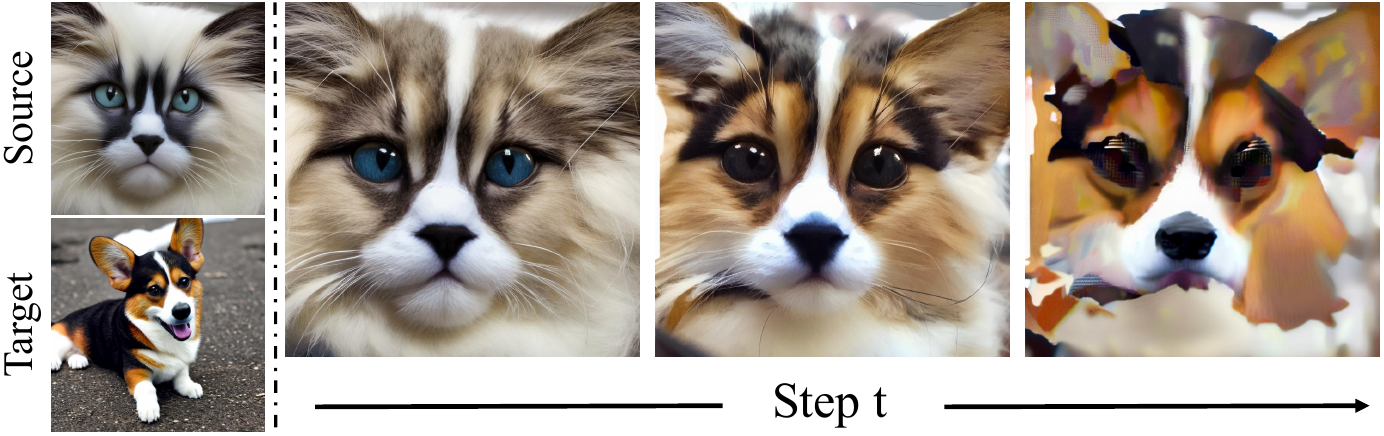}
    \caption{Visualization of the Ablation Study on End Step \( t \). The figure shows the gradual evolution of our generated images as the end step \( t \) increases.}
    \label{fig:ablation-step}
\end{figure}

\textbf{Ablation Study of End Step \( t \)}.  End Step \( t \) represents the step in our diffusion sampling process where the transfer is concluded. As shown in \cref{fig:ablation-step}, we find that the extent of appearance changes in our output images is closely correlated with the setting of the end step \( t \). This correlation is due to the progressive transfer process we adopt. When \( t \) is set too low, the transfer may lack sufficient appearance features, leading to incomplete or ineffective transfer. Conversely, setting \( t \) too high can result in over-transfer, incorporating excessive or redundant appearance features from the target. This suggests that the optimal appearance transfer occurs at the mid-stage of diffusion denoising.

\subsection{Limitation}
\begin{figure}[t]
    \centering
    \includegraphics[width=0.48\textwidth]{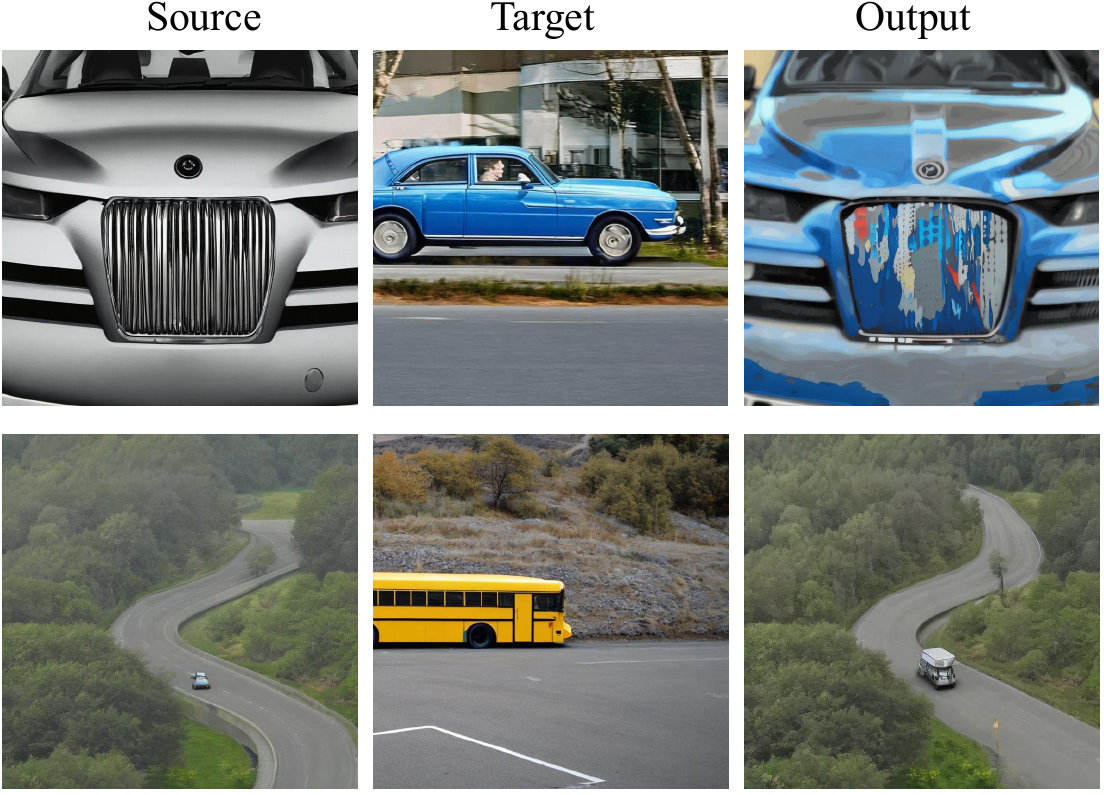}
    \caption{Illustration of failure cases of our method.}
    \label{fig:Limitation}
\end{figure}

Though achieving fine-grained appearance transfer, our method may fail under certain conditions, such as differing viewpoints and significant size discrepancies, as shown in~\cref{fig:Limitation}.
Due to differences in viewpoint, incorrect matches may occur between details in the source image (\eg the front grille of the car) and the target image, leading to the transfer of inappropriate appearance information. Additionally, if the main object in the source image is excessively small, the attention mask used in our method might be inaccurately positioned, resulting in erroneous appearance transfer.

\section{Conclusion}

In conclusion, this paper reports an important advancement in appearance transfer. We introduce a novel framework that effectively overcomes challenges in fine-grained transfer, utilizing the distinctive properties of the \( x_0 \) space within diffusion models. Our approach integrates semantic matching, appearance transfer, and latent deviation, ensuring the maintenance of structural integrity and information fidelity during fine-grained detail transfer.
A key aspect of our success is the seamless transition from \( x_0 \) space to latent space. This enables natural and accurate appearance transfers without necessitating extensive model training or fine-tuning. Comprehensive experiments across various categories and domains have thoroughly validated the versatility and robustness of our method. Our work enhances the diffusion-based models in precise and detailed image translation tasks.

As future directions, we plan to extend our focus to the broader context of general image transfer with an emphasis on fine-grained details. This expansion aims to further refine and apply our framework to a wider range of image translation scenarios, addressing more diverse and complex challenges in the field. Our future work will explore the potential of fine-grained transfer in various general image contexts, pushing the boundaries of what is achievable with current image translation technologies.

\clearpage

\section{Supplementary Details and Extended Analysis}

\subsection{Detailed Derivation of the Transformation Process}
We present a detailed derivation of our transformation process, focusing on the transition from the initial latent variable representation to the modified latent representation in the context of fine-grained appearance transfer.
Given an initial latent variable representation \( x_t \), it is composed of a data-informed component and a noise-driven term, formalized as:
\begin{equation}
    x_t = \sqrt{a_t} x_0 + \sqrt{1-a_t} \epsilon(x_t)
\end{equation}

where \( x_0 \) represents the structured data component and \( \epsilon(x_t) \) the noise term.
Applying a transformation in the \( x_0 \) space, denoted by \( \mathcal{T}(\cdot) \), we define a transformed latent variable \( x'_t \) as:
\begin{equation}
    x'_t = \sqrt{a_t} (x_0 + \mathcal{T}(x'_0, x_0)) + \sqrt{1-a_t} \epsilon(x_t) 
\label{eq.1}
\end{equation}

Here, \( \mathcal{T}(x'_0, x_0) \) denotes the space transformation function applied to \( x_0 \).
For the optimal latent representation \( x^*_t \), a linear combination of \( x_t \) and \( x'_t \) is considered:
\begin{equation}
    x^*_t = \lambda x_t + (1-\lambda) x'_t
\label{eq.2}
\end{equation}

where \( \lambda \) is the balancing coefficient.
The noise term \( \epsilon(x^*_t) \) is updated as a linear mixture:
\begin{equation}
    \epsilon(x^*_t) = \gamma \epsilon(x_t) + (1-\gamma) \epsilon(x'_t)
\label{eq.3}
\end{equation}

Here, \( \gamma \) is the mixing coefficient for the noise components.
For updating from \( x^*_t \) to \( x^*_{t-1} \), the following expression is derived:
\begin{equation}
    x^*_{t-1} = \sqrt{\frac{a_{t-1}}{a_t}} (x^*_t - \sqrt{1-a_t} \epsilon(x^*_t)) + \sqrt{1-a_{t-1}} \epsilon(x^*_t)
\label{eq.4}
\end{equation}

Substituting \cref{eq.1,eq.2,eq.3} into \cref{eq.4}, we obtain:
\begin{equation}
    x^*_{t-1} = \sqrt{\frac{a_{t-1}}{a_t}} (\lambda x_t + (1-\lambda) x'_t) + \kappa (\gamma \epsilon(x_t) + (1-\gamma) \epsilon(x'_t))
\label{eq.5}
\end{equation}

where \(\kappa\) is defined as \(\sqrt{1-a_t} - \sqrt{\frac{a_{t-1}(1-a_t)}{a_t}}\).
The final formulation of \cref{eq.5} can be further elaborated as:
\begin{equation}
\begin{split}
    x^*_{t-1} = & \sqrt{\alpha_{t-1}} (x_0 + (1-\lambda) \mathcal{T}(x'_0, x_0)) \\
                & + \sqrt{1-a_{t-1}} \epsilon(x^*_t) \\
                & + \sqrt{\frac{a_{t-1}(1-a_t)}{a_t}} (\epsilon(x_t) - \epsilon(x^*_t))
\end{split}
\label{eq.6}
\end{equation}

In \cref{eq.6}, the term \( (1-\lambda) \mathcal{T}(x'_0, x_0) \) indicates the shift within the \( x_0 \) space towards the target domain, while \( \epsilon(x^*_t) \) guides the direction towards \( x^*_t \). The term \( \epsilon(x_t) - \epsilon(x^*_t) \) adapts to the denoising process.

\subsection{Evaluation Metric}
Here we employ the CLIP score~\cite{radford2021learning} as a metric for evaluating semantic similarity in two distinct contexts: Text-to-Image (CLIP-T2I) and Image-to-Image (CLIP-I2I). The CLIP-T2I metric measures semantic similarity between the source prompt and the output image, whereas CLIP-I2I assesses semantic alignment between the source image and the output image.
As indicated in \cref{tab:more-clip}, both metrics highlight the superiority of our method. We specifically emphasize the use of CLIP-T2I for capturing semantic differences, aligning with the principles of general diffusion methods~\cite{chefer2023attend}.

\begin{table}[h]
    \centering
    \resizebox{0.7\linewidth}{!}{%
    \begin{tabular}{ccc}
        \toprule
        \textbf{Method} & CLIP-T2I~$\uparrow$ & CLIP-I2I~$\uparrow$ \\
        \midrule
        SplicingViT~\cite{tumanyan2022splicing} & 0.273 & 0.878 \\
        \midrule
        General I2I~\cite{cheng2023general} & 0.289 & 0.893 \\
        \midrule
        Exemplar~\cite{yang2023paint} & 0.268 & 0.872 \\
        \midrule
        Ours & \textbf{0.301} & \textbf{0.924} \\
        \bottomrule
    \end{tabular}
    }
    \caption{Comparison with Image-guided Transfer Methods on CLIP-T2I and CLIP-I2I Metric.}
    \label{tab:more-clip}
\end{table}

\subsection{Generation of Detailed Descriptions for Target Images}
\begin{figure}[h]
    \centering
    \includegraphics[width=0.48\textwidth]{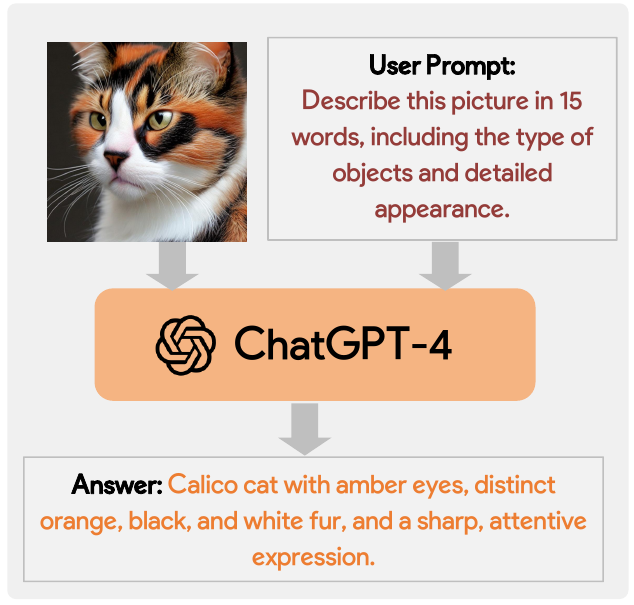}
    \caption{Illustration of generating detailed descriptions using ChatGPT-4 with multimodal input capabilities.}
    \label{fig:supp detail-text}
\end{figure}

In this section, we elucidate the process of generating detailed descriptions for target images as mentioned in Sec. 4 and illustrated in Fig. 6 of the main paper. Leveraging the multimodal input capabilities of online ChatGPT-4~\cite{OpenAI2023GPT4TR}, as demonstrated in \cref{fig:supp detail-text}, we input specific text along with target images to obtain descriptive outputs.

Our findings reveal that ChatGPT-4 demonstrates a robust capability for recognizing fine-grained categorical details and comprehending color and appearance aspects. However, it occasionally yields erroneous results in specific scenarios, such as heterochromia. To address these inaccuracies, we employed manual corrections post-hoc to ensure the precision of the textual outputs. This approach enabled us to refine the descriptions to more accurately match the visual characteristics of the target images.

\subsection{Data Set Collection}
In the main paper, we focus on several common animal species, including Horses, Cows, Cats, Dogs, Lions, Otters, Pandas, Foxes, and Tigers. Additionally, for generation purposes, we specified certain specific breeds within these categories to capture distinct appearance traits. For instance, in the case of dogs, we included breeds such as Corgis, Siberian Huskies, Border Collies, and Doberman Pinschers.

\subsection{Comparison of DIFT Matching Strategies}
\begin{figure}[h]
    \centering
    \includegraphics[width=0.48\textwidth]{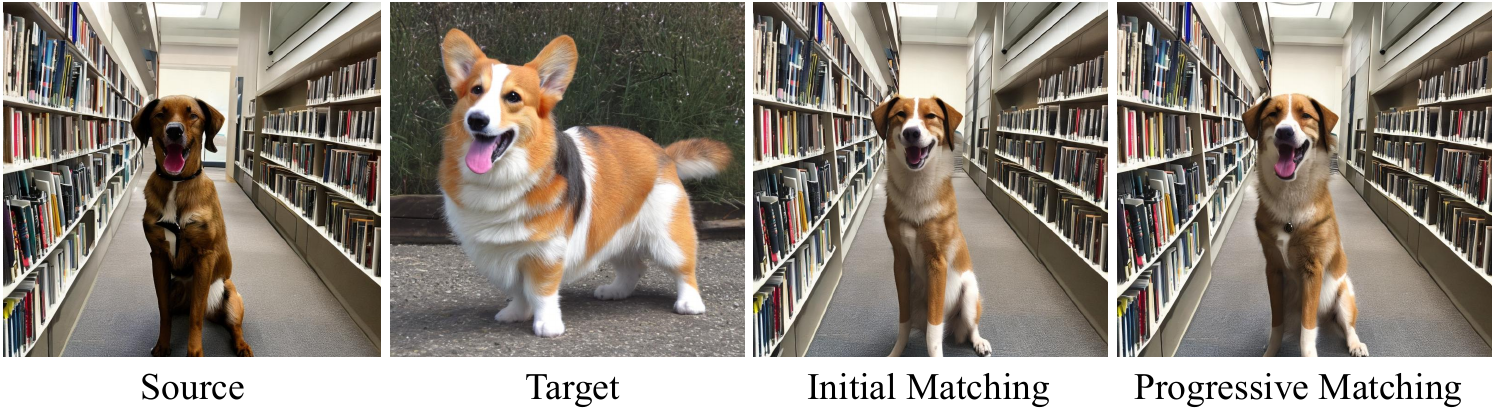}
    \caption{Illustration of Progressive and Initial DIFT Matching Strategies.}
    \label{fig:supp-matching}
\end{figure}

In the main paper, we utilize a progressive DIFT matching strategy within the \( x_0 \) space. Alternatively, DIFT matching can be implemented from the start by directly comparing the source and target images. As shown in~\cref{fig:supp-matching}, both approaches lead to relatively stable generative outcomes, as they both facilitate the establishment of relatively accurate matching relationships. While the progressive matching method emphasizes a transition in semantic granularity from coarse to fine, the initial matching approach offers the advantage of reducing the time spent in the matching phase.

\section{More Experimental Result}
\begin{figure*}[h]
    \centering
    \includegraphics[width=1.0\textwidth]{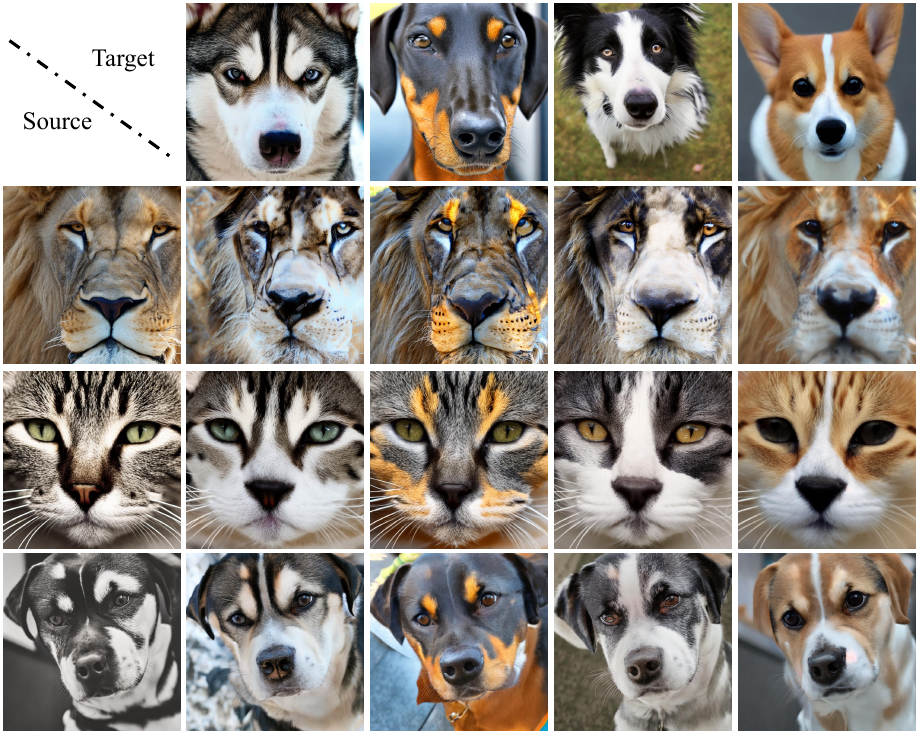}
    \caption{Demonstration of face appearance transfer between different animals. The target images represent dogs with distinct appearances, while the source images feature various animal species.}
    \label{fig:supp-animal}
\end{figure*}

\begin{figure*}[h]
    \centering
    \includegraphics[width=1.0\textwidth]{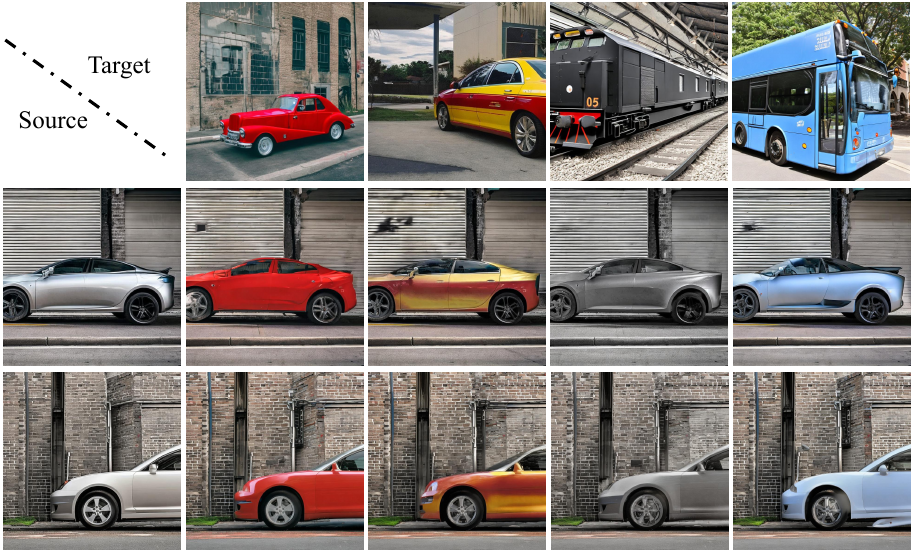}
    \caption{Appearance Transfer Results for Vehicles in Different Categories. The target images represent vehicles of various types, from left to right: a sedan, taxi, train, and bus.}
    \label{fig:supp-car}
\end{figure*}

\begin{figure*}[h]
    \centering
    \includegraphics[width=1.0\textwidth]{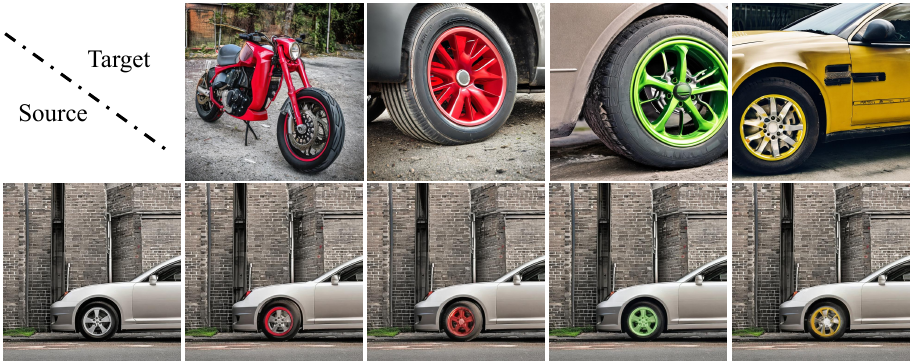}
    \caption{Fine-grained appearance transfer in automotive wheels. The target images showcase, from left to right: motorcycle wheels in the first column, followed by car wheels with varying appearance from the second to the fourth column.}
    \label{fig:supp-wheel}
\end{figure*}

In this section, we present additional experimental results. ~\cref{fig:supp-animal} shows the face appearance transfer between animals of similar and different categories.
~\cref{fig:supp-car} illustrates appearance transfers among different vehicles, while ~\cref{fig:supp-wheel} shows fine-grained transfers focusing on specific components, such as automotive wheel rims.

\newpage
\clearpage

{
    \small
    \bibliographystyle{ieeenat_fullname}
    \bibliography{main}
}


\end{document}